# Data-driven Method for Estimating Aircraft Mass from Quick Access Recorder using Aircraft Dynamics and Multilayer Perceptron Neural Network

Xinyu He, Fang He, Xinting Zhu, Lishuai Li[1]
Department of Systems Engineering and Engineering Management, City University of Hong Kong, Kowloon, Hong Kong

## Abstract

Accurate aircraft-mass estimation is critical to airlines from the safety-management and performance-optimization viewpoints. Overloading an aircraft with passengers and baggage might result in a safety hazard. In contrast, not fully utilizing an aircraft's payload-carrying capacity undermines its operational efficiency and airline profitability. However, accurate determination of the aircraft mass for each operating flight is not feasible because it is impractical to weigh each aircraft component, including the payload. The existing methods for aircraft-mass estimation are dependent on the aircraft- and engine-performance parameters, which are usually considered proprietary information. Moreover, the values of these parameters vary under different operating conditions while those of others might be subject to large estimation errors. This paper presents a data-driven method involving use of the quick access recorder (QAR)—a digital flight-data recorder—installed on all aircrafts to record the initial aircraft climb mass during each flight. The method requires users to select appropriate parameters among several thousand others recorded by the QAR using physical models. The selected data are subsequently processed and provided as input to a multilayer perceptron neural network for building the model for initial-climb aircraft-mass prediction. Thus, the proposed method offers the advantages of both the model-based and data-driven approaches for aircraft-mass estimation. Because this method does not explicitly rely on any aircraft or engine parameter, it is universally applicable to all aircraft types. In this study, the proposed method was applied to a set of Boeing 777-300ER aircrafts, the results of which demonstrated reasonable accuracy. Airlines can use this tool to better utilize aircraft's payload.

**Nomenclature**

| | |
|---|---|
| $T$ | Thrust |
| $D$ | Drag |
| $L$ | Lift |
| $m$ | Aircraft mass |
| $g$ | Gravitational acceleration |
| $V_a$ | True airspeed |
| $\theta$ | Pitch angle |
| $\psi$ | Roll angle |
| $\mu$ | True track angle |
| $\alpha$ | Angle of attack |
| $\gamma$ | Flight path angle |
| $q$ | Dynamic pressure |
| $S$ | Wing reference area |
| $M$ | Mach number |
| $a_x$ | Longitudinal acceleration |
| $a_y$ | Lateral acceleration |
| $a_z$ | Normal acceleration |
| $h$ | Altitude |
| $V_v$ | Vertical speed |

[1] lishuai.li@cityu.edu.hk

| | | | |
|---|---|---|---|
| $a_\theta$ | Pitch angle rate | $V_g$ | Ground speed |
| $C_L$ | Lift coefficient | $C_{L_0}$ | Zero lift coefficient |
| $C_{L_\alpha}$ | Lift slope coefficient | $C_{D_0}$ | Zero drag coefficient |
| $K$ | Drag polar constant | $Reg$ | Registration number |

## 1. Introduction

The aircraft mass is an important parameter for aircraft-performance analysis, trajectory prediction, etc (He et al., 2018). Moreover accurate aircraft-mass determination is necessary to obtain a better estimate of engine fuel consumption and cargo capacity, which are important from the viewpoint of profit maximization and ensuring flight safety. Thus, inaccurate aircraft-mass estimations can be considered a significant error source in all flight-operation-related calculations (Jackson et al., 1999).

Typically, airlines calculate the aircraft mass by adding the masses of its different components—empty aircraft, fuel, cargo, passengers, and crewmembers. However, it is impractical to weigh passengers and their carry-on baggage owing to privacy concerns. Accordingly, the airlines use rough estimates of these mass components to obtain an approximate aircraft mass. This estimated aircraft mass is logged into a flight-management system (FMS) by the pilots. An overestimation of the aircraft mass would result in it carrying less cargo compared to its rated capacity and more fuel than it would consume. This would not only reduce airline profitability but also result in more emissions than necessary. In contrast, underestimating the aircraft mass would result its overloading at takeoff. This might cause the aircraft to exceed its safety limits during certain in-flight maneuvers, thereby leading to a safety hazard. In reality, airlines carry more fuel than needed for each passenger, which not only pushes the cost of tickets up, but also means that far more emissions are produced than is needed. Thus, accurate aircraft-mass estimations are required for airlines to improve their operational strategies, such as maximizing efficiency while maintaining safety.

The flight data recorder (FDR) is installed in all aircrafts to record the values of in-flight operating parameters. The recorded data vary depending on the age and size of the aircraft, but it is a minimum requirement of all FDRs to record the values of five parameters—pressure altitude, indicated airspeed, magnetic heading, normal acceleration, and microphone keying. The FDRs installed in modern jet aircrafts record thousands of parameters to cover all aspects of aircraft operations and states. For example, the Airbus A380 aircraft records over 3,000 parameters on its FDR. However, data logging in the FDR requires significant preprocessing to derive parameter values from raw data. Meanwhile, the quick access recorder (QAR)—an airborne flight recorder—aims to provide quick and easy access to raw flight data. The parameters recorded by the QAR can be configured by the airline to be identical to or different from those recorded by the FDR. Figure 1 depicts the parameters recorded by the QAR during a flight of the aircraft considered in this study. It describes the observed trends in parameter values during different flight phases—takeoff, climb, cruise, descent, and approach to landing.

Because most parameters recorded by the FDR are considered classified information, only the airlines and authorized organizations have access to FDR data. Accordingly, the different methods for aircraft-mass estimation can be categorized into two types based on the use of open-source data (mainly flight-trajectory data) and FDR data.

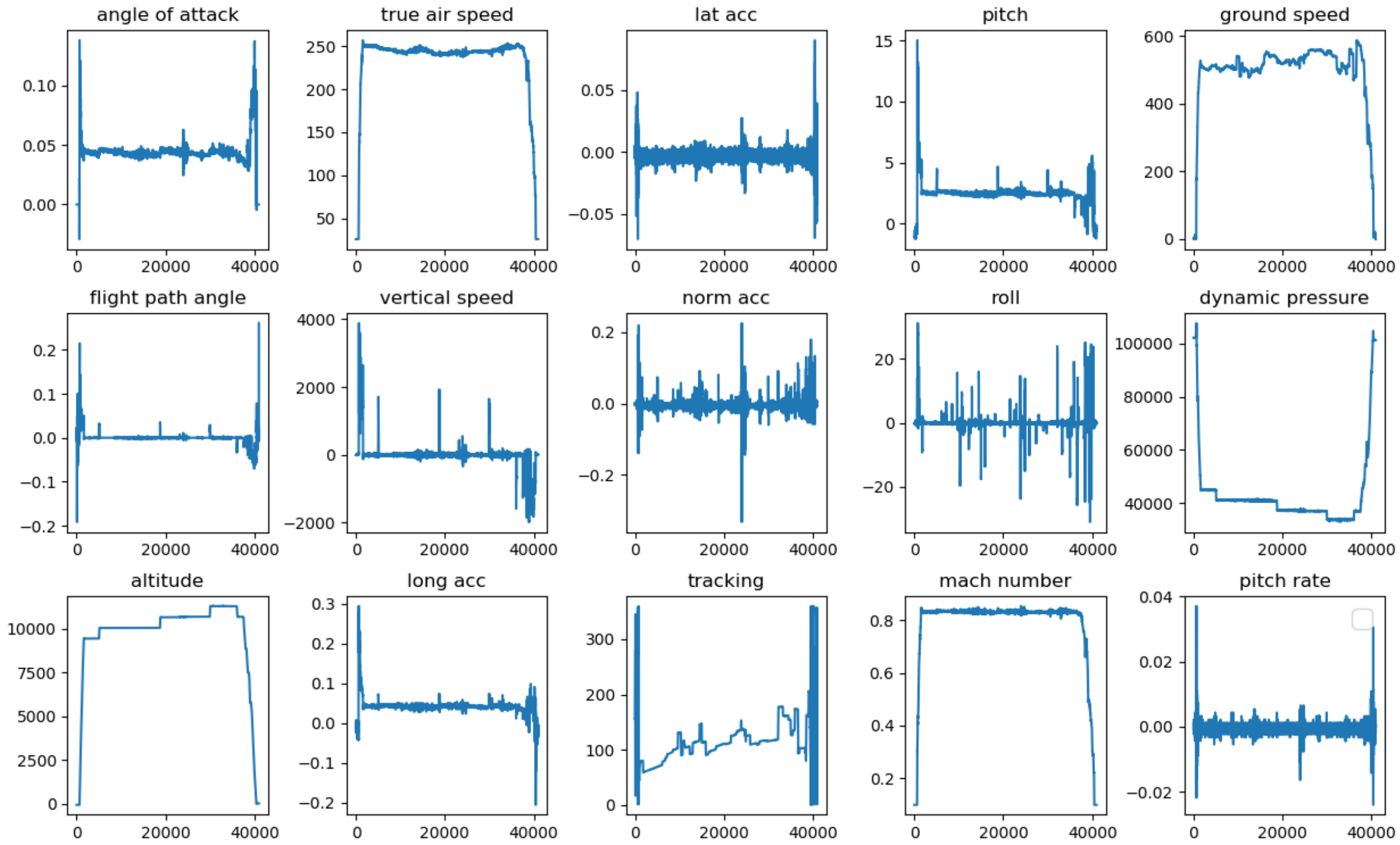

Figure 1: Trends in parameter values recorded as QAR data

The information regarding aircraft mass is rarely available with the air-traffic management (ATM) research community and air-traffic controllers. Thus, several methods based on the automatic dependent surveillance-broadcast (ADS-B) and radar track data have been developed to estimate the aircraft mass using flight-trajectory data. Schultz et al. (2012) proposed an adaptive radar-track-data-based method for aircraft-mass estimation during the climb phase to improve trajectory predictions. Their method dynamically adjusts the modeled aircraft mass by bridging the gap between the observed and predicted energy rates obtained from the track data and aircraft model, respectively. Similarly, Alligier et al. (2012, 2013) used the quasi-Newton algorithm to learn the thrust profile of aircrafts based on past trajectory records by minimizing the error between the predicted and observed energy rates. Their proposed approach could estimate the aircraft mass using the least-squares method based on a few operating points lying on the past trajectories as well as the learnt thrust profile. In addition to radar data, they used weather information to determine the wind and air temperatures for estimation. Further, Alligier et al. (2014) proposed a variation of their previous method by additionally considering fuel-consumption information. Subsequently, Alligier et al. (2015) proposed a completely different approach to determine the aircraft mass from a set of sample trajectories. In this approach, the missing aircraft mass is replaced by an adjusted mass that best fits the energy rate.

Several machine-learning-based regression algorithms, such as linear regression with forward selection, ridge regression, principal component regression, single-layer neural network, and gradient boost regression, have been used to predict the true aircraft mass. Using ADS-B data and physical kinetic models, Sun et al. (2016) proposed two least-squares-approximation-based analytical methods to estimate the takeoff aircraft mass. In the first method (Sun et al., 2017), meteorological data are combined with the ground speed in ADS-B to approximate the true airspeed. The method first calculates the aircraft mass using different methods, including with fuel-flow models, during different flight phases. Subsequently, these mass calculations are combined with the prior knowledge of the initial aircraft-mass probability distribution to yield the maximum a posteriori estimation based on a Bayesian inference approach. Sun et al. (2018) reports the investigation of the variations observed in the obtained results owing to dependent factors, such as prior distribution, thrust, and wind. Moreover, the results were validated against data recorded during 50 test flights of a Cessna Citation II aircraft. The validation results revealed a mean absolute error of 4.3% in the predicted mass. In a latter study, Sun et al. (2019) included wind and temperature data from the enhanced Mode-S dataset as additional observations alongside ADS-B data. They proposed a stochastic recursive Bayesian approach that employed a regularized sample importance re-sampling particle filter in conjunction with a nonlinear state space model. This approach could eliminate noise from observations as well as determined an acceptable noise limit to obtain an accurate aircraft-mass estimate. Lee and Chatterji (2010) estimated aircraft takeoff mass based on the approximation of each individual mass component and aircraft-performance data.

Three factors introduce errors in aircraft-mass estimations performed using trajectory data. The first corresponds to the unavailability of important flight parameters in the trajectory dataset. These missing parameters need to be substituted by other parameters; for example, flight path angle is used in the absence of the angle of attack. Second, because the aircraft mass is closely linked to thrust in flight-dynamic evaluations, any uncertainty in the engine thrust could result in a large difference in the estimated mass. Most existing methods estimate the aircraft mass under the maximum thrust profile assumption (Bartel and Young, 2008), which is not always hold in real-world flight operations. In addition, aircraft mass varies as a function of fuel burn when considering an entire flight from takeoff to landing. For mass derived from phases following the initial climb, the aircraft fuel consumption must be considered when determining aircraft initial mass. However, fuel consumption can only be evaluated by extracting a fuel-burn model from trajectory data. These models introduce additional errors in the predicted result, despite their good approximation capability. Most extant studies have exclusively used simulated data or small samples of data recorded onboard an actual flight. In addition, all above-described methods involve the use of model-based approaches with the following three disadvantages. First, these approaches are heavily dependent on the availability of precise parameter values, and any inaccuracies in parameter values yields distorted results. Second, expert domain knowledge is required to develop an appropriate model. Lastly, aircraft systems are dynamically complex and highly nonlinear. Accordingly, the model-based approaches are require to solve multiple higher-order equation to achieve good accuracy; however, linearized approximation models are preferred for use in practical applications.

Since only airline operators and small groups of authorized researchers have access to FDR data, very few methods use FDR data to estimate the aircraft mass. The FDR records the ground truth of the takeoff mass of an aircraft. Unlike prior methods that use flight-trajectory data, Chati and Balakrishnan (2017, 2018) employed the Gaussian process regression (GPR) method to determine the operational takeoff weight of an aircraft using the data recorded during the takeoff ground roll. Their proposed approach uses the physical understanding of aircraft dynamics during the takeoff ground roll to select appropriate variables for building the GPR model. As reported, this method achieved a mean absolute error of 3.6% in the predicted aircraft takeoff mass. This is nearly 35% smaller compared to that incurred by models based on the aircraft noise and performance databases. However, such approaches rely heavily on the availability of accurate aircraft- and engine-performance, such as aerodynamic coefficients and thrust profiles, which are proprietary information, and therefore, difficult to obtain. Even if this information could be made available from the aircraft manufacturer or via use of open-source reference data, such as BADA3 (Nuic, 2010), errors may still be introduced in the prediction results owing to variations in the flight Mach number, Reynolds number, and aircraft configurations in real-world scenarios.

In data-driven approaches, the physical model is substituted by a statistical machine-learning algorithm. In addition, data-driven methods do not require complex dynamic modeling, and they can work satisfactorily when supplied with only a few parameter values recorded by the FDR. The superior performance of these methods could be attributed to the better learning capacity of the statistical machine- and deep-learning models. However, the lower interpretability of data-driven methods compared to model-based approaches is a major limitation. Therefore, there exists a motivation to combine the advantages of the model-based and data-driven methods. The major challenge here lies in combining the physical and statistical models used within the model-based and data-driven approaches, respectively.

To bridge the gap between the model-based and data-driven methods, this paper presents a method that uses QAR data to estimate the initial-climb aircraft mass. The aircraft mass recorded in the FMS is included as the ground truth in the QAR dataset. Because most airlines routinely collect and analyze such QAR data as part of the flight operational-quality assurance or flight data monitoring programs, our proposed method is suitable and easy to use in actual flight operations. Because flight parameters that are not available in the radar or ADS-B data can now be used, the proposed method achieves higher accuracy compared to prior model-based approaches. The key idea here is to use a physical model to select parameters from the QAR dataset and represent the aircraft mass as an implicit function of these parameters. Subsequently, the said function can be evaluated using statistical machine-learning methods. The selected parameters represent the bridge between the physical and statistical models. In the proposed method, a dynamic model was used to select appropriate parameter values, which were subsequently cleaned and smoothened to remove any abnormalities and noise interference from the recorded dataset. Finally, this pre-processed dataset is used to learn the approximation function. It is reported that the multilayer perceptron neural network (MLPNN) can approximate any function (Hornik, 1991; Hornik et al., 1989; Leshno et al., 1993). Thus, MLPNN was used in this study to approximate the aircraft-mass-estimation function. In contrast to prior methods, the proposed approach was validated using a large actual dataset comprising QAR data recorded on aircrafts flying over different routes over several years. The contributions of this study are twofold.

(1) The proposed uses a simple dynamic model instead of the sophisticated ones used in model-based methods. Thus, no expert domain knowledge is required to use this approach. Additionally, the deep-learning-based model is more interpretable compared to the data-driven methods.
(2) Because the target equation does not depend on the engine thrust, knowledge of aircraft-specific information, such as aerodynamic coefficients and engine thrust rating, is not required.

The remainder of this paper is structured as follows. Section 2 illustrates the methodology followed in this study, including the selection of useful flight parameters using a physical model, data preprocessing, and a description of the MLPNN model's architecture and the corresponding input/output. Section 3 describes the evaluation and testing of the proposed method when applied to datasets obtained from a single and multiple aircrafts. Finally, Section 4 lists major conclusions drawn from this study and identifies the scope for future research.

## 2. Methodology

The method proposed in this paper uses QAR data to estimate the initial mass of an aircraft in climb, and Figure 2 depicts the corresponding workflow. In this method, we first determine the useful parameters among several thousand others recorded by the QAR via use of a physical model. Subsequently, the selected data are processed to eliminate abnormal in-flight parameter values followed by data smoothening. Finally, the processed data are used to train the MLPNN model for aircraft-mass estimation.

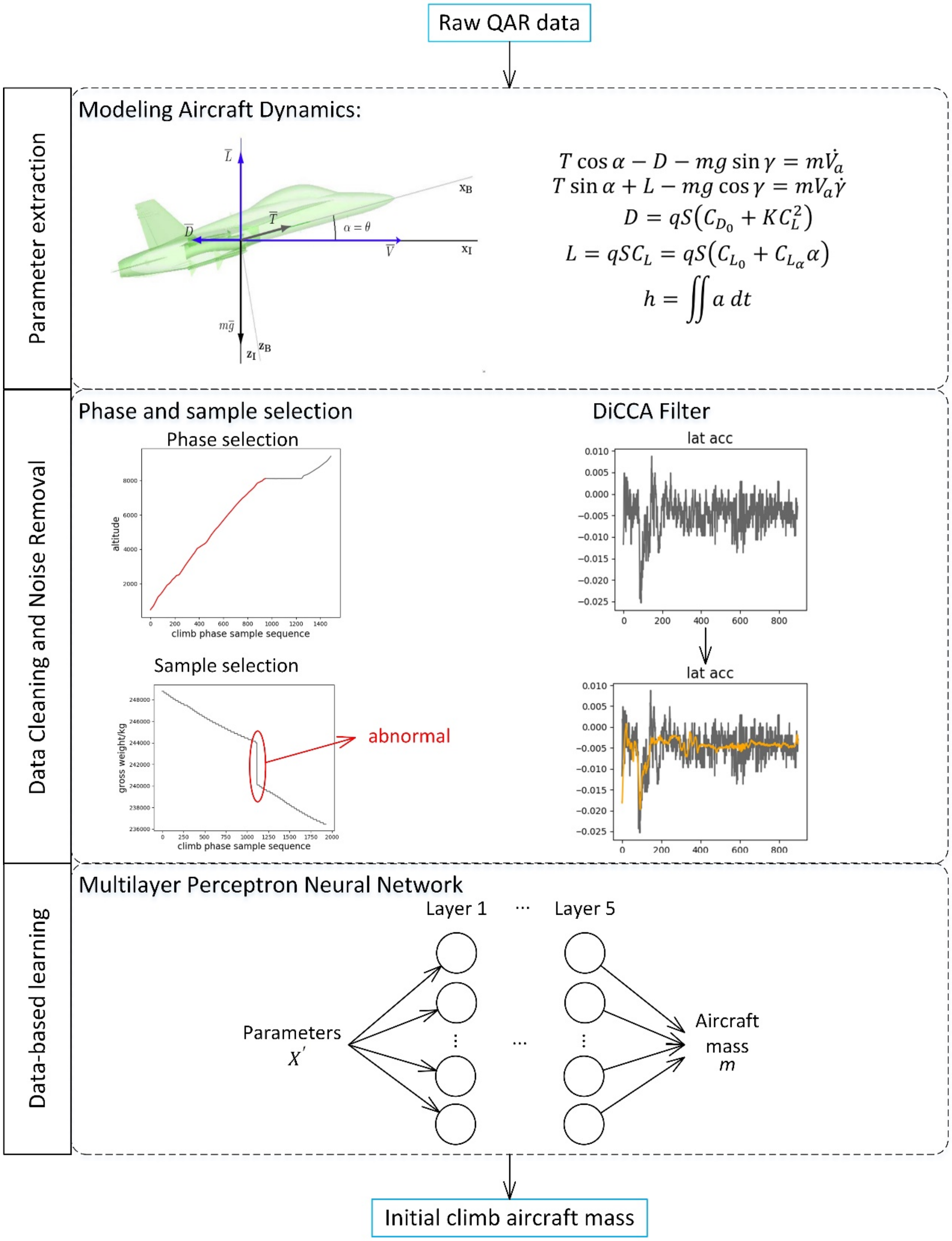

Figure 2: Flowchart of proposed aircraft-mass-estimation method

## 2.1 Parameter Extraction

This section describes the process for selection of the required flight parameters from the set of all parameters recorded by the QAR through dynamic flight equations. The raw QAR dataset contains thousands of parameters recorded at different sampling rates and encoded in specific binary formats. To make such data usable, the raw binary data must be decoded into usable values and subsequently resampled at a fixed interval. Lastly, we need to select useful parameters from thousands of parameters. Instead of the statistical parameter-selection algorithms employed in most data-driven methods (for example, the LASSO algorithm(Tibshirani, 1996)), we used the dynamic flight equations to determine the useful parameters to be extracted from the QAR dataset. This approach was favored because it does not require much expert knowledge and lends higher interpretability to the proposed method. Accordingly, we modeled the aircraft dynamics, the cumulative effects of which were considered. The said approach can be described as follows:

A non-rolling aircraft at any instant satisfies the following aircraft-dynamics equations.

$$T\cos\alpha - D - mg\sin\gamma = m\dot{V}_a \tag{1}$$

$$T\sin\alpha + L - mg\cos\gamma = mV_a\dot{\gamma} \tag{2}$$

Eliminating the thrust $T$ from the above equations, we get

$$mg\cos\gamma + mV_a\dot{\gamma} - L - \left(D + mg\sin\gamma + m\dot{V}_a\right)\tan\alpha = 0 \tag{3}$$

Based on aerodynamic considerations (Anderson Jr, 2010), the lift $L$ and drag $D$ forces can be expressed as

$$L = qSC_L = qS\left(C_{L_0} + C_{L_\alpha}\alpha\right) \tag{4}$$

$$D = qS\left(C_{D_0} + KC_L^2\right) \tag{5}$$

It is noteworthy that the values of the coefficients in Eqs. (4) and (5) do not remain constant during the climb phase. By substituting Eqs. (4) and (5) in Eq. (3), we get

$$\begin{aligned} &q\left(SC_{L_0}\right) + q\alpha\left(SC_{L_\alpha}\right) + q\tan\alpha\left(SC_{D_0} + SKC_{L_0}^2\right) + q\alpha\tan\alpha\left(2SKC_{L_0}C_{L_\alpha}\right) \\ &\quad + q\alpha^2\tan\alpha\left(SKC_{L_\alpha}^2\right) + m\left(g\sin\gamma\tan\alpha + \dot{V}_a\tan\alpha - V_a\dot{\gamma} - g\cos\gamma\right) = 0 \end{aligned} \tag{6}$$

It must be noted that Eq. (6) holds exclusively for non-rolling flights. However, this requirement may not always be satisfied in practice, and other non-linear relationships may hold between these parameters and the aircraft mass. Therefore, deducing the value of $m$ directly from Eq. (6) may introduce large errors. Nonetheless, it is useful to represent the implicit relationship between $m$ and the above-mentioned flight parameters, because it helps one identify which flight parameters must be considered to determine $m$.

For Eq. (6), during the climb phase, although some parameters are readily available from the QAR dataset, others need to be derived from the same. For example, the acceleration due to gravity $g$ acting on an aircraft at a given altitude $h$ can be approximated considering the below relationship.

$$g = g_0\left(\frac{R_e}{R_e + h}\right)^2 \tag{7}$$

In the above equation, $R_e = 6.3781 \times 10^6\ m$ denote the earth's mean radius, and $g_0 = 9.80665\ m/s^2$ denotes the gravitational acceleration on the earth's surface. The true airspeed rate $\dot{V}$ denotes the aircraft's acceleration, which can be substituted by Cartesian components $a_x, a_y, a_z$ in the QAR dataset. Likewise, the flight-path angle rate $\dot{\gamma}$ can be substituted by the pitch angle rate $a_\theta$. Because Eq. (6) is applicable exclusively to non-rolling flights and coefficients therein vary with Mach number $M$, the values of $\psi$ and $M$ must be selected. Flight parameters, such as the vertical speed $V_v$, ground speed $V_g$, pitch angle $\theta$, and track angle $\mu$, must also be introduced to reduce the uncertainties in the above equations. Thus, the below expression can be considered.

$$X = \left[\alpha, \gamma, \theta, \psi, \mu, V_a, V_v, V_g, a_\theta, h, a_x, a_y, a_z, M, q\right] \tag{8}$$

Accordingly, for each flight $i$, the corresponding aircraft mass $m_i$ can be expressed as

$$m_i = f_1(X_i) \tag{9}$$

Since registration numbers are used to identify different aircrafts, all flights operated using a given aircraft have same registration number and they can be expressed using the same function $f_1$ . This is because the aircraft- and engine-performance parameters (e.g., aerodynamic characteristics and thrust rating) remain nearly identical.

To this point in the dynamic analysis of the aircraft in climb, we have considered the influence of parameters at every instant of the aircraft motion. Although, uncertainties and noise interference may exist in the recorded dataset at each instant, the cumulative uncertainties and noise are somewhat lower. Thus, the performance of the proposed model can be improved by considering the cumulative influence of the above-described parameters. In accordance with Newton's second law of motion, the mass of an object can be represented using the force and acceleration acting on it. The typical cumulative effect of the force and acceleration acting on an object is its displacement expressed as

$$h = \iint a\, dt, \tag{10}$$

where $h$ denotes the displacement, which corresponds to the aircraft altitude in this case, and $a$ denotes the acceleration. Therefore, the time interval $\Delta t$ and altitude $\Delta h$ gained during $\Delta t$ can be considered parameters of the aircraft's persistent state.

$$m_i = f_2(\Delta h_i, \Delta t_i) \tag{11}$$

Thus, the aircraft mass can be represented as a function of both the instantaneous and cumulative parameters as follows.

$$m_i = f_1(X_i) + f_2(\Delta h_i, \Delta t_i) = f(X_i, \Delta h_i, \Delta t_i) \tag{12}$$

Moreover, to make $f$ applicable to all aircrafts, the flight-registration number can be considered an independent input. Thus, the mass of any given aircraft can be represented using a single expression given by

$$m = f(X, \Delta h, \Delta t, Reg) \tag{13}$$

The only in-flight reduction in aircraft mass occurs owing to fuel consumption by the engines. Thus, the instantaneous gross mass $m_j$ of an aircraft can be evaluated as the difference between its initial takeoff mass $m_I$ and mass of the fuel consumed $m_{f,j}$ in the interim; that is,

$$m_j = m_I - m_{f,j} \tag{14}$$

The mass $m_{f,j}$ at instant $j$ during the climb phase can be evaluated using the fuel-consumption rates of the left ($\dot{m}_l$) and right ($\dot{m}_r$) engines from the takeoff instant $k$ to instant $j$ as recorded by the QAR. That is,

$$m_{f,j} = \int_{t_k}^{t_j} (\dot{m}_l + \dot{m}_r)\, dt \tag{15}$$

Therefore, the final expression for the initial climb mass $m_I$ of an aircraft can be written as

$$m_I = f(X, \Delta h, \Delta t, Reg) + m_f \tag{16}$$

Table 1 lists all necessary parameters from the QAR dataset required to evaluate $m_I$ using Eq. (16).

Table 1 : Parameters extracted from raw QAR data

| Name | Symbol |
|---|---|
| Aircraft mass (gross weight) | $m$ |
| Fuel flow rates of left engine | $\dot{m}_l$ |
| Fuel flow rates of right engine | $\dot{m}_r$ |
| Altitude | $h$ |
| Dynamic pressure | $q$ |
| Angle of attack | $\alpha$ |
| Flight path angle | $\gamma$ |
| Pitch angle rate | $a_\theta$ |
| True airspeed | $V_a$ |
| Longitudinal acceleration | $a_x$ |
| Lateral acceleration | $a_y$ |
| Normal acceleration | $a_z$ |
| Roll angle | $\psi$ |
| Mach number | $M$ |
| Registration number | $Reg$ |
| Vertical speed | $V_v$ |
| Ground speed | $V_g$ |
| True track angle | $\mu$ |
| Pitch angle | $\theta$ |
| Altitude gain | $\Delta h$ |
| Sample length | $\Delta t$ |

## 2.2 Data cleansing and noise removal

The parameters selected as described in the previous section must be processed prior to providing as input to the proposed model for aircraft-mass estimation. During processing, we first select parameter-value samples corresponding to the longest climb segment and check if the calculated aircraft mass demonstrates large fluctuations. Subsequently, the data smoothened to eliminate noise from selected parameters.

Unlike the cruise phase, wherein most flight parameters remain nearly steady and the effective sample size is small, the values of parameters listed in Table 1 demonstrate large variations with time during the climb phase. Thus, the effective sample size is large and the function $f$ is easily learnt by the estimation model. Most flights perform multiple climb maneuvers to reach their cruise altitude. Figure 3 Figure 3depicts a typical altitude-gain curve for an aircraft. Because the aircraft dynamics vary significantly during the climb phase, input parameters for the proposed model were extracted from the longest continuous climb segment.

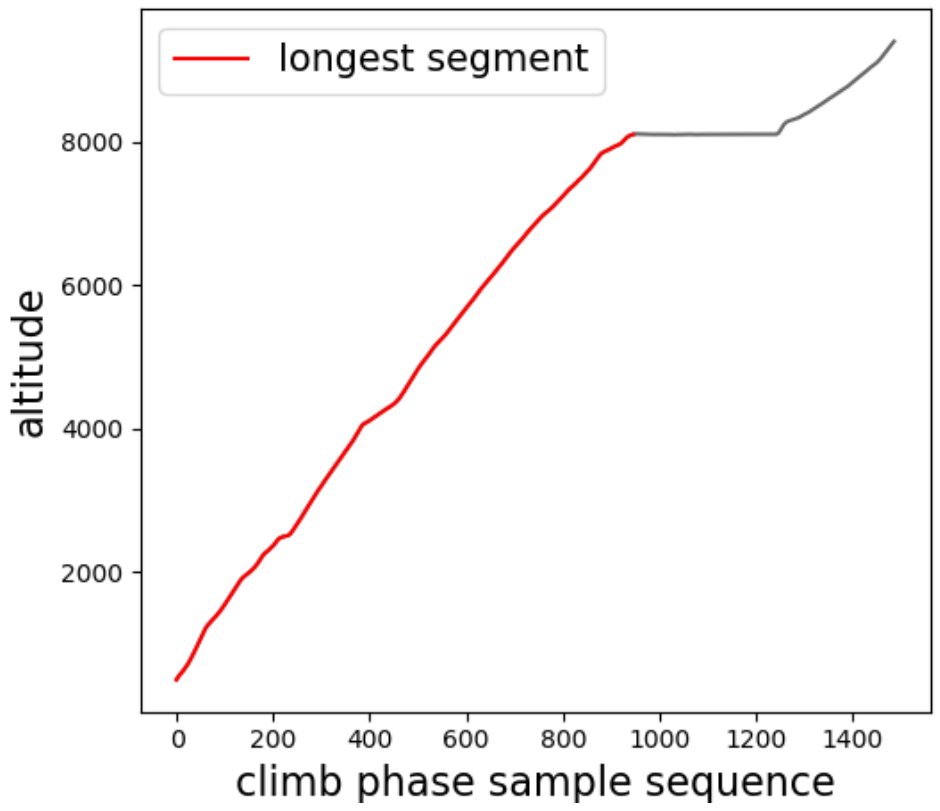

Figure 3: Altitude-gain curve for aircraft in climb phase

The aircraft-mass values recorded by the QAR equipped with the proposed MLPNN model might demonstrate large fluctuations owing incorrect recording and errors incurred during the decoding of raw binary data. Thus, data cleaning is required to ensure high accuracy. In this study, we considered the data-cleaning operation was performed based on three rules. First, the recorded masses of the aircraft $m_j$ and fuel consumed $m_{f,j}$ must remain positive; that is, $\forall\, j, m_j > 0\;\&\; m_{f,j} > 0$. Second, the values of $m_j$ and $m_{f,j}$ must monotonically decrease and increase, respectively. In addition, according to Eq. (14), the sum $m_j + m_{f,j}$ must remain constant. In this study, we considered $\max_j(m_j + m_{f,j}) - \min_j(m_j + m_{f,j}) < L$, and $L$ = 300 kg. Abnormal QAR data samples that do not satisfy these rules, they stand eliminated or are substituted by the neighboring sample value. The replacement is performed if the fluctuation appears transient. However, the sample values are eliminated if the fluctuations do not disappear post mutation.

As noise exists in parameters selected from the QAR dataset, especially in the acceleration data recorded by the sensors, data smoothening becomes imperative. In this study, we chose the dynamic-inner canonical correlation and causality analysis (DiCCA) algorithm (Dong and Qin, 2018) over filters, such as the Kalman filter, to remove noise from sampled data. Figure 4 and Figure 5 depict results obtained using the DiCCA algorithm. The input parameters to DiCCA contain 15 dimensions (additional parameters that remain constant during a given flight are not considered here), and the number of dynamic latent variables (DLVs) was set to 14. The DiCCA algorithm extracts the principal features (lowest-frequency signals) from sampled data as DLVs. This is accomplished by first maximizing the correlation between DLVs and their predictions. This is followed by principal removal from sampled data, and the residual data is used to generate more DLVs in an iterative manner. It is noteworthy that data can be reconstructed from DLV predictions. The higher the DLV value, the greater is the number of high-frequency signals. This can be confirmed from Figure 4, wherein DLV 1 corresponds to a single smooth curve whereas DLV 14 contains the most fluctuations. Thus, the DiCCA algorithm offers a convenient means to eliminate high-frequency noise signals from sampled data by controlling the DLV count. However, it must be noted that too many DLVs may result in retention of substantial noise while a low DLV count might result in the omission of useful information. As depicted in Figure 5, the trends concerning parameters $a_y$ and $a_\theta$, which are the most contaminated by high-frequency noise, become much smoother post application of the DiCCA algorithm. Before proving as input to the MLPNN model, the training data are z-score normalized followed by determination of their mean and standard deviation (SD) values. These mean and SD values are later used to normalize the test data.

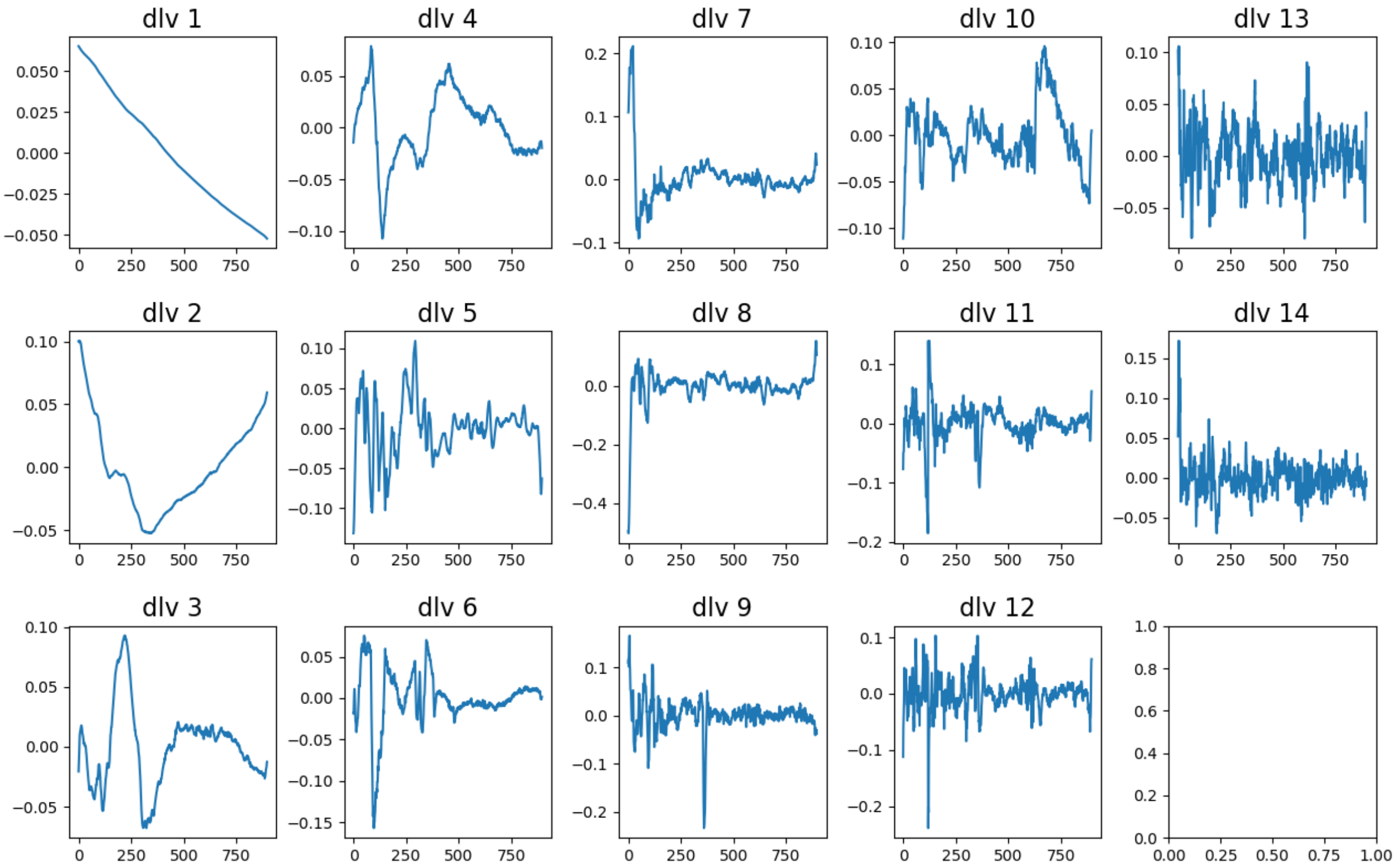

Figure 4: Dynamic latent variables obtained via application of DiCCA to sample flight dataset

## 2.3 Regression using MLPNN

This section explains the regression method employed in the proposed aircraft mass estimation model. During regression, the QAR data were divided into three datasets—training, validation, and test. Sixty percent of all flights were randomly selected for building the training dataset. Meanwhile, 20% flights were randomly selected to form the validation dataset for selection from a group of candidate models, and the remaining 20% flights constituted the test dataset for evaluating the proposed-model performance.

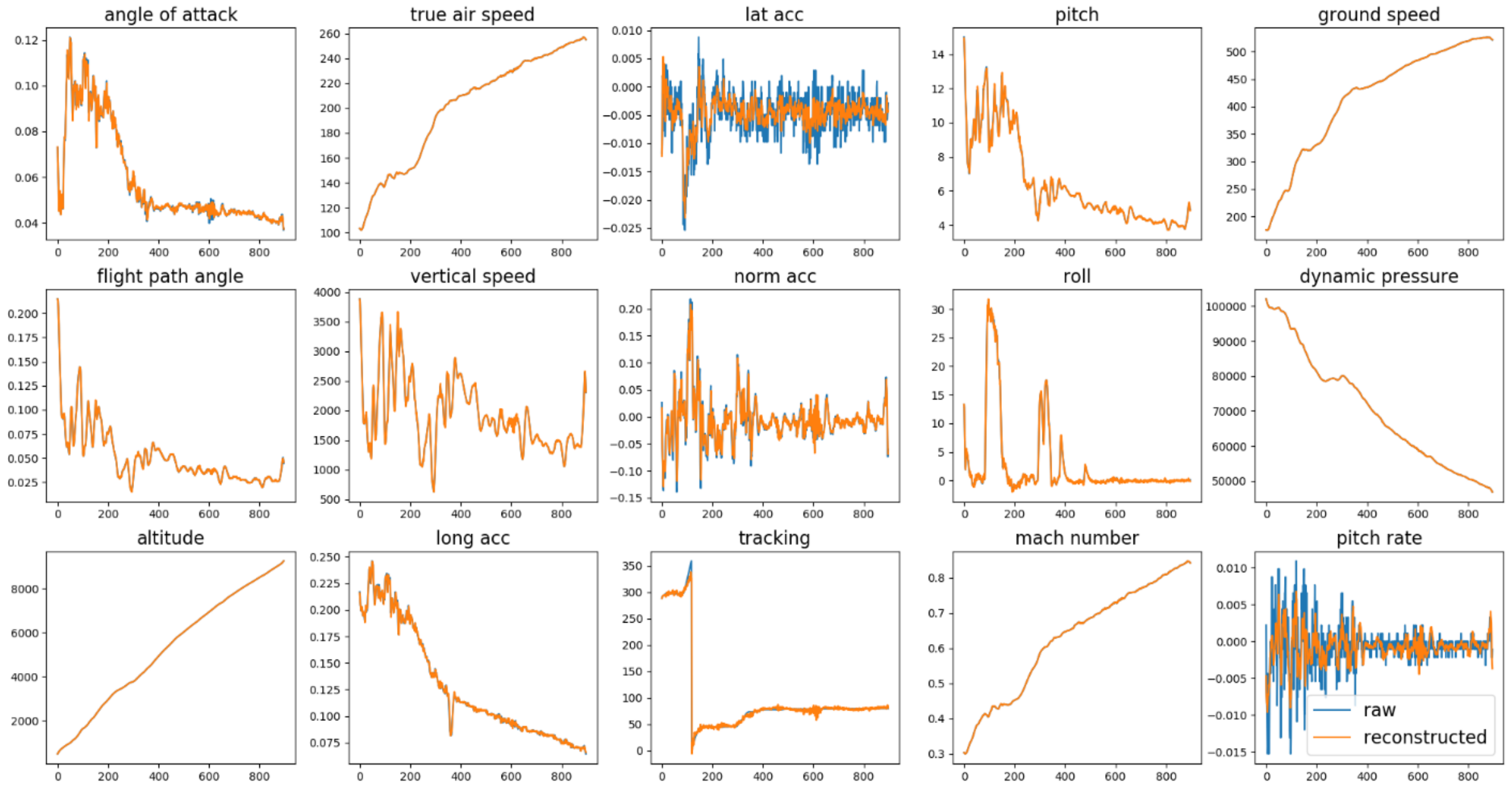

Figure 5: Flight parameters smoothened using DiCCA algorithm

It is difficult to evaluate $m$ explicitly using Eq. (13) owing to the requirement to satisfy several conditions and solve multiple equations. Accordingly, regression techniques should be used to evaluate the function in Eq. (13) using available data. The commonly used statistical regression methods include the support vector regression (SVR), GPR, decision-tree regression (DTR), and linear regression. However, performing SVR and GPR involves the use of kernels, the size of which equals the square of the number of samples. This makes it hard to scale to datasets with more than a couple of 10000 samples. Thus, they are not efficient for large datasets owing to the large memory requirements and fit time complexity. Recently, the application of deep-learning techniques has attracted significant research attention owing to their superior performance compared to statistical regression methods. In theory, the MLPNN-based approach can approximate any function (Hornik et al., 1989). Thus, an MLPNN-based model has been used in this study to determine $m$ using the QAR dataset. The results obtained have been compared against those obtained using other methods.

Using the MLPNN model, Eq. (13) can be expressed as

$$m = f(X, \Delta h, \Delta t, Reg) \approx g(X'; \phi) \tag{17}$$

where $X' = [X, \Delta h, \Delta t, Reg]$, $g$ denotes the MLPNN model, and $\phi$ denotes the parameter of $g$. Correspondingly, the equation to estimate the initial aircraft mass can be expressed as

$$m_I \approx g(X'; \phi) + m_f \tag{18}$$

The tendency of nonlinear activation functions—*sigmoid* and *tanh*—to become saturated when supplied with a large input makes them unsuitable for use in approximation problems. Accordingly, *Relu* was used in this study. It is noteworthy that although the deep-architecture *Relu* neural networks are more efficient compared to their shallow counterparts (Yarotsky, 2017), they are significantly more difficult to train. Thus, the MLPNN model used in this study was designed to be neither too deep nor shallow. Batch normalization (BN) was considered to accelerate the model training.

## 3. Evaluation and Testing

This section describes the application of the proposed method to real-world scenarios. As already mentioned, the actual QAR dataset recorded on-board a Boeing 777-300 ER aircraft was used in this study. The relative error between the aircraft mass estimated using the proposed method and that recorded by the QAR was calculated. The proposed model was trained using data recorded during different flights of the same aircraft, albeit the number of flights was small. In addition, the training dataset contained parameter values recorded during several other flights of different aircrafts. The results reveal that the more the number of flights considered for preparing the training dataset the better is the observed performance and generalization of the proposed model.

### 3.1 Data Description

The dataset comprised parameter values recoded during 3,480 flights of the Boeing 777-300 ER aircraft fleet flying over different routes between 2016 and 2018. This dataset was obtained from an international airline company. The fleet comprised 19 aircraft; that is 19 different flight registration numbers, as depicted in Figure 6(A). As can be seen, the flights with registration numbers 1 and 19 completed the maximum (237) and minimum (161) number of flights, respectively. Figure 6(B) depicts the mass distribution of the aircraft fleet, and the maximum and minimum values of the QAR-recorded initial climb masses equal 350,942 kg and 227,409 kg, respectively. Similar to a Gaussian function with an upper bound, the mass distribution conforms to a bell shape. This is because of the limited payload capacity of these aircrafts, and the resulting maximum gross-weight regulations for this aircraft type.

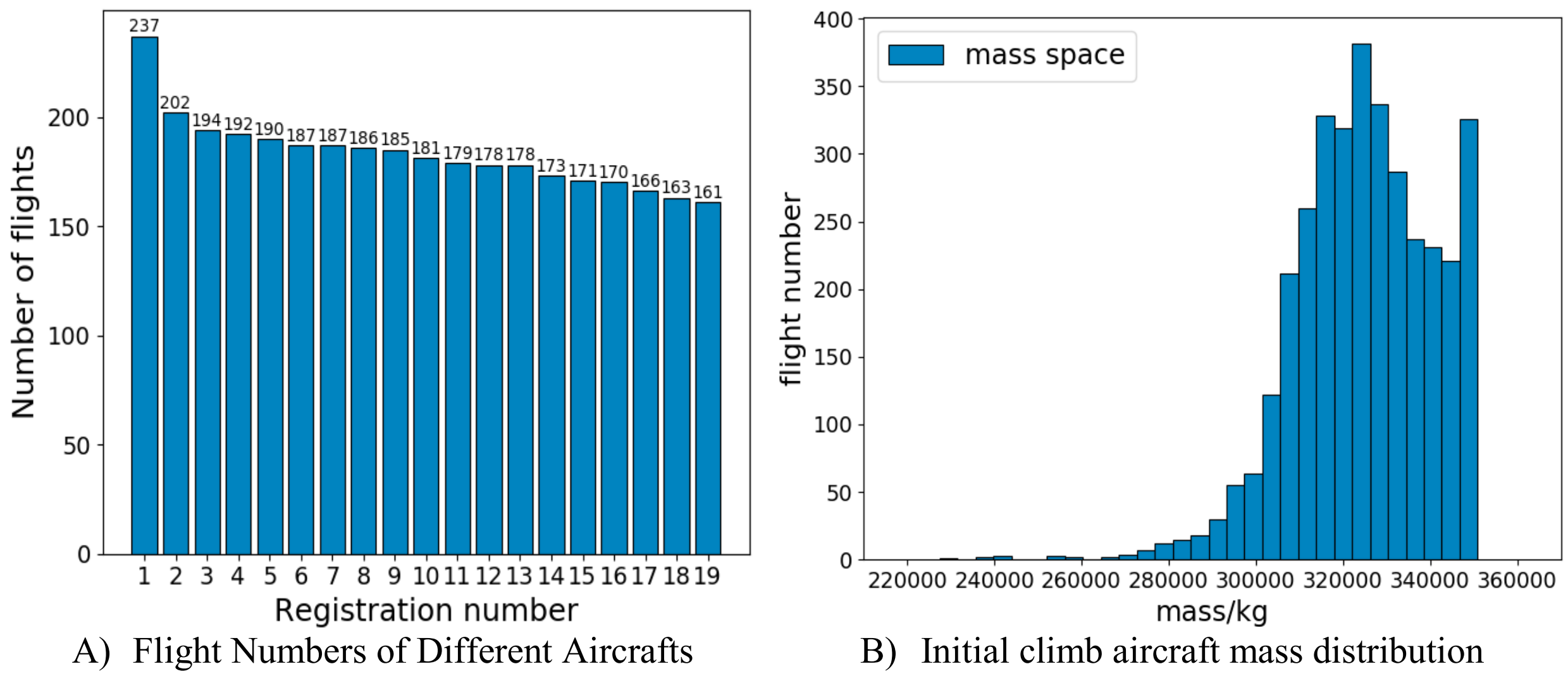

A) Flight Numbers of Different Aircrafts B) Initial climb aircraft mass distribution

Figure 6: Initial climb aircraft mass distribution in dataset

### 3.2 Evaluation and testing of MLPNN model for Boeing 777-300 ER

In this study, QAR data recorded during 190 flights of a single aircraft (registration no: 5) were used to test the proposed method. Subsequently, the number of samples was increased to include the data recorded on all aircrafts (3,480 flights). The recoded data were divided into the training (60%), validation (20%), and test (20%) datasets. As already mentioned, results obtained using the proposed MLPNN model were compared against those obtained using other algorithms—DTR and ridge regression (RR). The metrics used to compare the evaluation results included the mean absolute percentage error (MAPE), normalized root-mean-square deviation (NRMSD) and $R^2$. Accordingly, the model demonstrating lower MAPE and NRMSD as well as high $R^2$ values were preferred.

- MAPE indicates the $L_1$-norm accuracy of the prediction results. It corresponds to the mean of the absolute relative prediction error given by

$$MAPE = \frac{1}{n^*}\sum_{i=1}^{n^*}\left|\frac{m_i - \hat{m}_i}{m_i}\right|, \tag{19}$$

  where $m_i$ and $\hat{m}_i$ denote the QAR-recorded and predicted masses of an aircraft during flight $i$; $n^*$ denotes the number of flights considered in the test dataset.

- NRMSD indicates the $L_2$-norm accuracy of the prediction result. It denotes the squared difference between the observed and predicted values and can be expressed as

$$NRMSD = \frac{\sqrt{\frac{1}{n^*}\sum_{i=1}^{n^*}(m_i - \hat{m}_i)^2}}{m_{max} - m_{min}}, \tag{20}$$

  where $m_{max}$ and $m_{min}$ denote the maximum and minimum values of the aircraft mass in the considered aircraft fleet.

- The $R^2$ score indicates the goodness of fit of the proposed model, thereby providing a measure of how well the proposed model is likely to predict unknown values. It is given by

$$R^2 = 1 - \frac{\sum_{i=1}^{n^*}(m_i - \hat{m}_i)^2}{\sum_{i=1}^{n^*}\left(m_i - \frac{1}{n^*}\sum_{i=1}^{n^*} m_i\right)^2}, \tag{21}$$

Table 2 compares the parameter settings for the MLPNN and DTR algorithms. The values of the hyper-parameters, such as the hidden-layer depth, layer size, and L2 penalty parameters, pertaining to the MLPNN model were tuned in accordance with the speed and performance. In addition, the values of hyper-parameters concerning the DTR algorithm were also tuned to overcome overfitting.

Table 2: MLPNN- and DTR-based model parameter settings

| Regression method | Model details | Selected parameters | Training range |
|---|---|---|---|
| MLPNN | Hidden-layers depth | 5 layers (Dense + BN) | [3,9] |
| | Layer size | 80 perceptrons per layer | [32,100] |
| | Activation function | 'Linear' for last layer, 'Relu' for the rest | - |
| | Error function | Mean square error | - |
| | L2 penalty parameters | 0.01 | [0.001,0.05] |
| | Solver for optimization | 'Adam' | - |
| DTR | Maximum tree depth | 10 for one aircraft, 15 for others | [10, 25] |
| | Minimum sample count to split an internal node | 10 for one aircraft, 20 for others | [5, 25] |
| | Minimum sample count to consider a leaf node | 5 for one aircraft, 10 for others | [3, 15] |
| | Complexity parameter for minimum cost-complexity pruning | 0.2 | [0, 0.4] |

Table 3 presents a comparison between results obtained using the three algorithms. Because the number of samples contained in the dataset obtained from a single aircraft is small, the MLPNN model demonstrates good performance on the training set, albeit the prediction performance is sub-optimal (MAPE increases from 0.11% to 1.17%; i.e., 10 times the error). This implies that the proposed model is overfitted. Meanwhile, the DTR and RR results reveal a more severe overfit and negligible overfit, respectively. With increase in sample count (when considering the entire aircraft fleet), the observed overfitting is alleviated for both MLPNN and DTR. That is, MLPNN achieves good performance when applied to the test dataset, and the corresponding performance of DTR is improved as well.

Although hundreds of samples are selected from each flight and the total sample count obtained for a given aircraft exceeds 100,000, the effective sample count is not very large. Because the initial climb aircraft mass for a given flight performed by an aircraft would remain nearly unchanged, it can be considered representative of a single sample. Thus, although the aircraft-mass estimation obtained using the MLPNN model is overfitted for a single aircraft, the model is more generalized for all aircrafts.

Figure 7 depicts a comparison between the predicted masses of all aircrafts obtained using the MLPNN, DTR, and RR methods. As can be seen, the MLPNN model achieves better accuracy and lower variance compared to the other methods. Figure 8 depicts the relative errors incurred when applying the MLPNN model to the test dataset containing samples from all aircrafts. As observed, the relative error for most flights lies in the [-2.0%, 2.0%] range.

Table 3: Initial-climb aircraft-mass estimation: performance metrics for MLPNN, RR, and DTR methods

| Aircraft number | Dataset (number of flights) | Metrics | MLPNN | RR | DTR |
|---|---|---|---|---|---|
| Single aircraft | Training (114) & Validation (38) | MAPE (%) | **0.11** | 1.70 | 0.47 |
| | | NRMSD | **0.005** | 0.089 | 0.030 |
| | | $R^2$ | **0.999** | 0.749 | 0.972 |
| | Test (38) | MAPE (%) | **1.17** | 1.90 | 2.46 |
| | | NRMSD | **0.084** | 0.118 | 0.193 |
| | | $R^2$ | **0.869** | 0.738 | 0.295 |
| All aircrafts | Training (2088) & Validation (696) | MAPE (%) | **0.30** | 1.75 | 0.63 |
| | | NRMSD | **0.010** | 0.063 | 0.025 |
| | | $R^2$ | **0.994** | 0.769 | 0.964 |
| | Test (696) | MAPE (%) | **0.61** | 1.72 | 1.48 |
| | | NRMSD | **0.033** | 0.088 | 0.082 |
| | | $R^2$ | **0.970** | 0.782 | 0.808 |

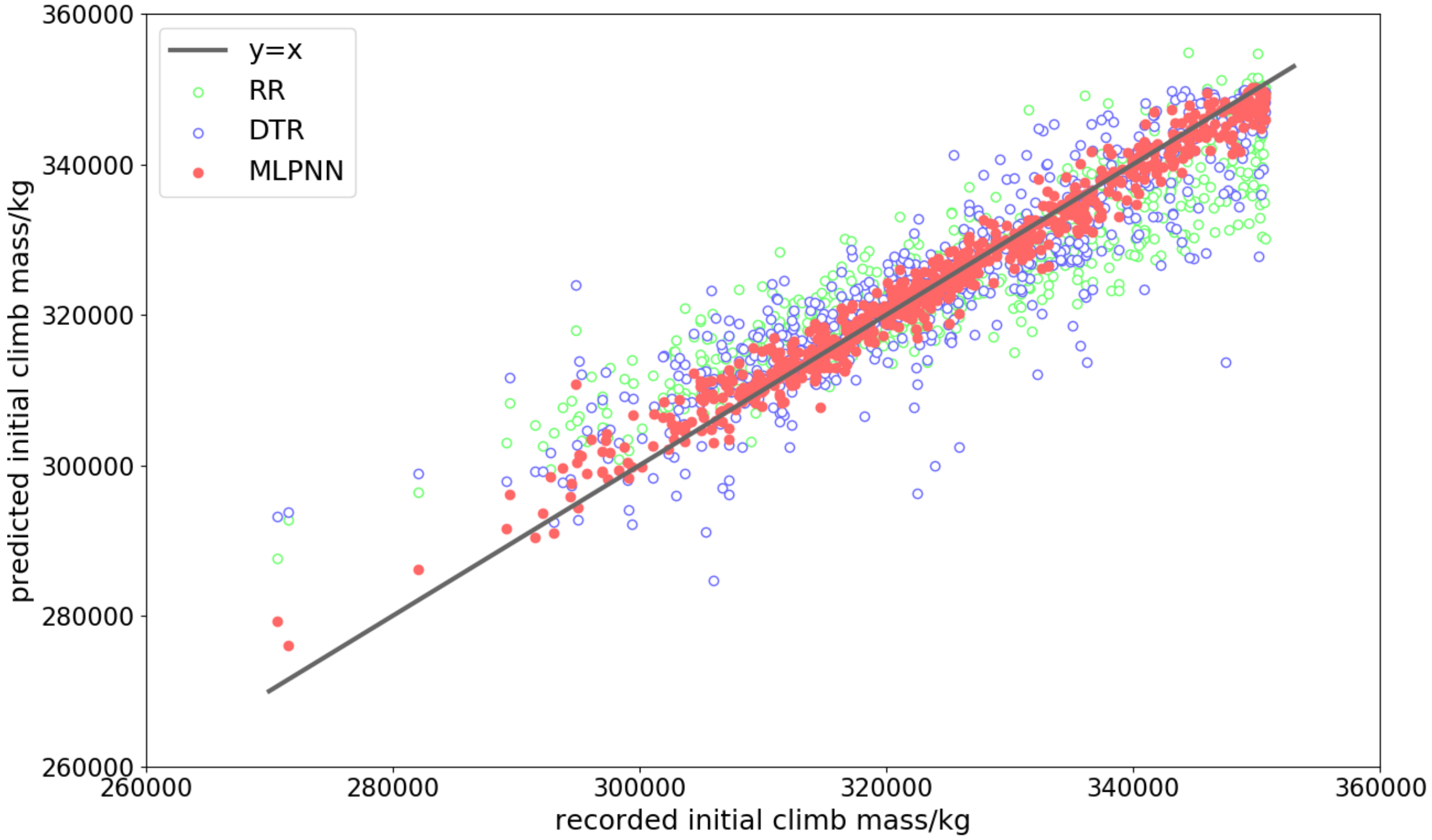

Figure 7: Mass estimation obtained by applying the three methods on the test dataset containing samples from all flights of B777-300 ER aircrafts

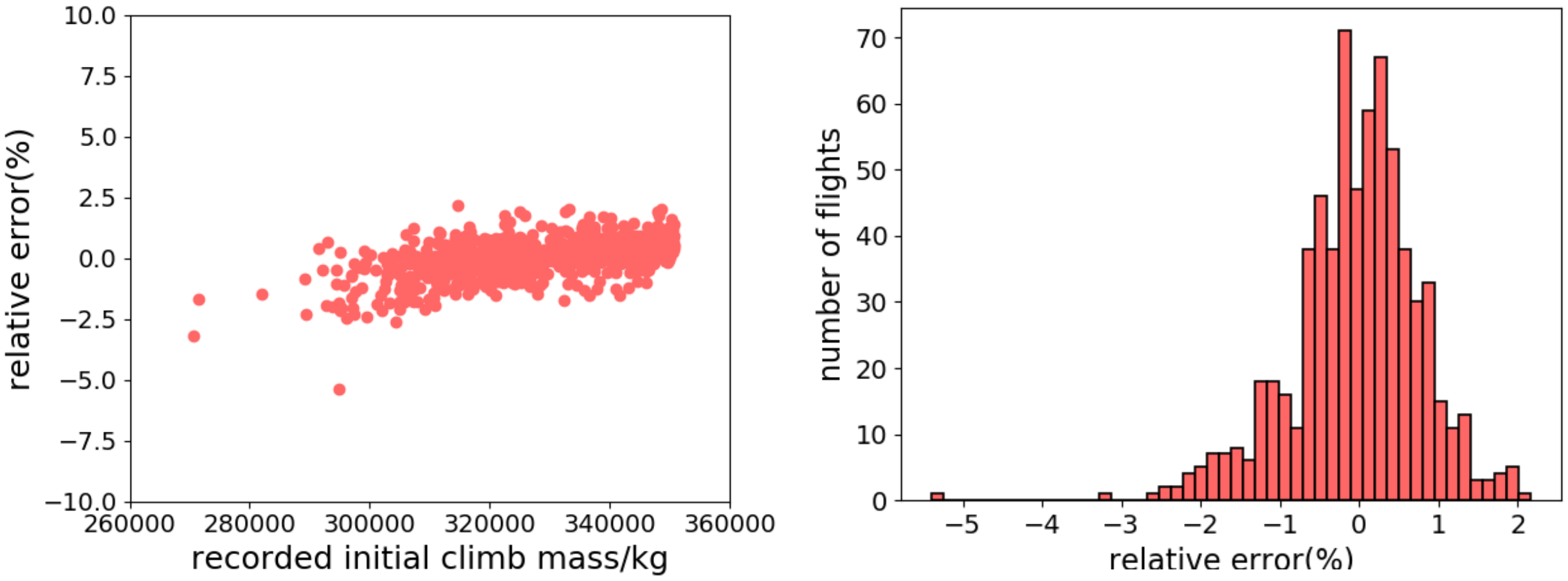

A) Relative error in mass estimation B) Error distribution over flight count

Figure 8: Relative error observed when applying MLPNN method to test dataset

The examination of Figure 8(B) reveals the existence of a flight with relative error exceeding -5%. The parameter values recorded during this flight were separately analyzed to investigate the cause of the large prediction error. Figure 9 depicts the selected parameter trends for this flight. As explained in Section 2, the selection of flight parameters to estimate the initial-climb aircraft mass was performed under the assumption of non-rolling flights. However, as can be seen in Figure 9, the rolling motion of this aircraft is characterized by frequent fluctuations, thereby complicating its true dynamic motion. The use of the MLPNN model demonstrates an improved prediction accuracy owing to the absence of error-inducing approximations and substitutions in physical models. However, if the roll attitude of an aircraft changes frequently, the resulting parameter-value fluctuations in the prediction model would induce uncertainties and inaccuracies. Unlike other flights, the rolling motion pertaining to this flight demonstrates greater fluctuations. Thus, the prediction results for this flight incur the largest absolute percentage error.

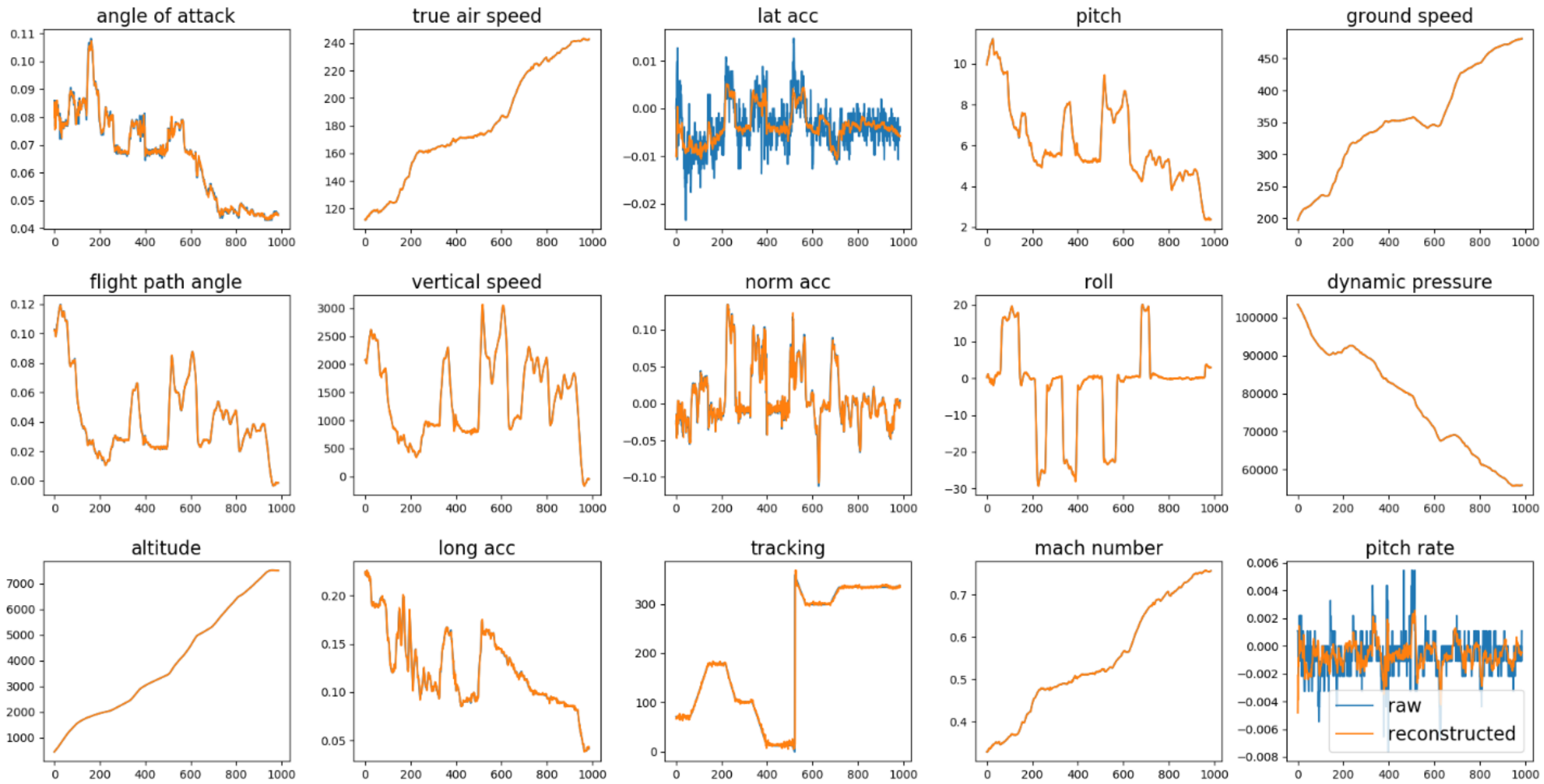

Figure 9: Selected parameters for flight demonstrating largest absolute percentage error in predicted mass

## 4. Conclusions

This paper presents a method to obtain an accurate estimate of the aircraft mass using QAR data. The proposed method acts as a bridge between the physical model-based and statistical approaches. It supplies flight parameters obtained from a physical model as input to the statistical model. The selected parameters neither depend on the engine thrust nor include any aircraft-specific information pertaining to its geometry and/or aerodynamic coefficients. Data pre-processing and noise removal are performed to ensure high quality flight-parameter data and aircraft-mass labels. Finally, the multilayer perceptron neural network has been selected as the statistical model to perform regression. In this study, the proposed method was first tested using QAR data recorded during 696 flights and subsequently applied to corresponding datasets recoded during 3,480 flights. As observed, the results obtained reveal superior accuracy, lower errors, and better generalization capability of the proposed method compared to state-of-the-art regression models. In future endeavors, the authors intend to extend the applicability of this method to other flight phases that may involve different aircraft dynamic-motion scenarios and selection of different flight parameters. A major drawback of this approach is the large flight-data volume required to realize model generalization and can be future work.

## Acknowledgment

This work was supported by the Hong Kong Research Grants Council (Project No. 11215119, 11209717 and 21202716).